# A SURVEY ON PHRASE STRUCTURE LEARNING METHODS FOR TEXT CLASSIFICATION

Reshma Prasad[1] and Mary Priya Sebastian[2]

[1]Department of Computer Science and Engineering, Rajagiri School of Engg. & Technology, Kakkanad, Kochi, Kerala

[2]Department of Computer Science and Engineering, Rajagiri School of Engg. & Technology, Kakkanad, Kochi, Kerala

*ABSTRACT*

*Text classification is a task of automatic classification of text into one of the predefined categories. The problem of text classification has been widely studied in different communities like natural language processing, data mining and information retrieval. Text classification is an important constituent in many information management tasks like topic identification, spam filtering, email routing, language identification, genre classification, readability assessment etc. The performance of text classification improves notably when phrase patterns are used. The use of phrase patterns helps in capturing non-local behaviours and thus helps in the improvement of text classification task. Phrase structure extraction is the first step to continue with the phrase pattern identification. In this survey, detailed study of phrase structure learning methods have been carried out. This will enable future work in several NLP tasks, which uses syntactic information from phrase structure like grammar checkers, question answering, information extraction, machine translation, text classification. The paper also provides different levels of classification and detailed comparison of the phrase structure learning methods.*

*KEYWORDS*

*Text classification, Phrase structure, Phrase patterns, Natural Language Processing (NLP)*

## 1. INTRODUCTION

Text classification or categorization includes automatic classification of documents or texts into predefined categories. Different application of text classification includes spam filtering, email routing, language identification, genre classification, readability assessment etc. There are different methods for text classification which includes decision trees [2], rule based classifiers [5], SVM classifiers [7], neural network classifiers [4], bayesian classifiers [3] and nearest neighbour classifiers [1]. Text classification can be improved if phrase patterns are used in the classification task and phrase pattern identification progresses with already extracted phrases.

Phrase structure is the grammatical arrangement of words in a sentence. The words in a sentence are not arranged in just any order, but language has constraints on word order. Words are organized into phrases, groupings of words that are clumped as a unit and a sentence can be modeled as a set of phrases. Syntactic knowledge can be modeled by phrase structure. Phrase structure is the backbone of many models of syntax of natural language. It can be a powerful way to express sophisticated relations among the words in a sentence. NLP activities like grammar checkers, question answering, information extraction, machine translation, text classification uses syntax information form phrase structure.

The advantage of using phrase patterns for text classification is that phrases helps to identify long distant dependencies, the structure can support distant relationships between words. This method provides flexibilities of modeling local word-reordering or grouping   which is the main problem of free order languages. Also, phrase extraction brings in some semantic value, which is suitable when NLP activities like text classification, or machine translation is considered. Another advantage of the phrase pattern identification is that it filters words occurring frequently in isolation that does not have much weight towards classification. The different methods of phrase structure learning or extraction have to be studied, classified and compared.

## 2. SURVEYED TECHNIQUES

The goal of the phrase structure extraction is to automatically extract phrase structures from a given corpus. The different techniques surveyed here are based on phrase structure learning. Many methods of phrase structure learning have already been developed for languages like English, Chinese, German, Japanese, Swedish etc. Phrase extraction techniques based on both bilingual and monolingual corpus are discussed in this paper. Different methods surveyed here are:

### 2.1. Basic N-gram based approach

N-gram based approach is a statistical approach which includes application of n-gram models to obtain phrases. William B. Cavnar and John M. Trenkle proposed an approach which extracted phrases using n-gram model and the phrases thus obtained are used for text categorization [6]. Gulila Altenbek, Ruina Sun used N-gram models for phrase structure extraction from unannotated monolingual corpus [31]. Bigram and trigram models are applied to extract phrases from the corpus. The monolingual corpus is roughly segmented and N-gram model is applied followed by a normalization process. Equation (2) represents the probability [31].

$$P\left(\frac{W_n}{W_{n-1}}\right) = \frac{C(W_{n-1}W_n)}{C(W_{n-1})} \qquad (2)$$

Accuracy is measured in terms of number of phrases correct to total phrases extracted. The accuracy  is measured around 51%, which is low. Among the two models, bigram model has more accuracy than trigram [31].

### 2.2. Rule based method

Ramshow and Marcus used transformation rule based learning for extracting the noun phrases[9]. Gulila Altenbek, Ruina Sun used rule based method [31] for noun phrase extraction from monolingual corpus. The method is a non-statistical approach which uses annotated monolingual corpus. The approach is based on the basic rules of the target monolingual language; therefore developing a rule set for the corresponding language is a necessary condition. The phrases are extracted according to the rules defined, the corpus is searched for a matched rule and the phrases thus found are extracted.

Accuracy is measured in terms of number of phrases correct to total phrases extracted. The accuracy for rule based approach is found to be around 80% while that of N-gram based approach is found to be around 51%.

## 2.3. Word alignment based method

Word alignment based method is a statistical method. Phrase alignments are learned from a corpus that has been word aligned. The basic idea is to align the parallel corpus in both directions and to take an intersection so that an alignment matrix is generated. The alignment points in the alignment matrix are expanded based on different heuristics.

Franz Josef Och, Christoph Tillman and Hermann Ney developed an improved alignment model [11] which uses alignment templates. An alignment template is a triple (F,E,A) where A is an alignment matrix with binary values. The template describes the alignment between source class sequence F and target class sequence E. The initial step of the alignment template approach is to align the parallel corpus in the two translation directions, source to target and target to source. Expectation maximization algorithm is applied in both directions to obtain two alignment vectors. The two alignment vectors are combined to form the alignment matrix A. Iteratively checking and adding neighboring links extend the alignment. All the consistent phrase pairs of the training corpus are determined by checking if the source phrase words are aligned only to target phrase words.

The advantage of this approach is the fully automatic learning using bilingual training corpus [11]. The disadvantage is that it selects all templates without checking whether it is good or bad [11]. The measuring score used is word error rate (WER), position independent error rate (PER) and subjective sentence error rate (SSER). In terms of efficiency, the error rate decreased to about 6% than baseline.

Philip Koehn, Franz Josef Och and Daniel Marcu modified the alignment template approach later [16]. The heuristics for expansion in the alignment template approach is modified by permitting diagonal neighborhood in the expansion stage [5]. Giza++ toolkit is used for word alignment. Lexical weighting and maximum phrase length scores are applied to the model. Top performance is obtained when the phrase length is three. The method shows better performance when lexical weighting score is applied. The method has an improvement of about 0.01 BLEU score than alignment template approach.

## 2.4. Phrase alignment based method

Phrase alignment based method is another statistical approach in which phrases are extracted from the phrase alignment using phrase-based joint probability.

Daniel Marcu and William Wong developed a phrase based joint probability model [14]. The model captures simultaneous generation of source and target sentences in a parallel corpus rather than alignment between them. In this method, each sentence pair in our corpus is generated by the idea of generation of a bag of concepts (each concept is a phrase pair) and the bag of concepts can be arranged linearly to obtain source and target sentences [14]. The initial step of the method is to find high frequency n-grams and t-distribution table initialization. Expectation maximization learning on the Viterbi alignment is then applied iteratively which yields joint probability distribution. The performance of the phrase-based method has an average improvement of about 0.02 BLEU score than word alignment based method.

A modification of the base model has been proposed by Philip Koehn, Franz Josef Och and Daniel Marcu [16]. The base model is modified by marginalizing the joint probabilities to conditional probability [16]. The performance of the method is high when phrases are of length three.

International Journal on Natural Language Computing (IJNLC) Vol. 3, No.2, April 2014

### 2.5. Syntactic approach

Philip Koehn, Franz Josef Och and Daniel Marcu proposed a syntactic method [16], which involves parsing of the sentences in bilingual parallel corpus. In this method, a syntactic phrase is defined as a sequence of words which is covered in a single subtree of a syntactic parse tree [15]. The first step of the method is to word align the parallel corpus. Both side of the corpus is parsed using syntactic parsers. For consistent phrase pairs, it is checked whether both the phrases are subtrees in the parse trees generated. The measuring score used is BLEU score. A BLEU score of 0.243 is obtained when efficiency is measured [16].

### 2.6.. Mutual Information based method

Ying Zhang, Stephan Vogel and Alex Waibel developed an integrated phrase segmentation and alignment algorithm [17] for statistical machine translation which uses Mutual Information (MI). The algorithm segments sentences into phrases and thus can be used as a phrase extraction technique. In this method, an initial word alignment or initial segmentation on the monolingual text is not required [17]. The phrases are identified by similarity of point wise mutual information and thus the sentences are segmented into phrases. A two dimensional matrix is constructed to represent the source and target sentence pairs where value of each cell corresponds the point wise mutual information score between source and target words. From the matrix, phrase pair with high MI value is selected and it is expanded to rectangle regions such that the expanded region has a similar MI value. The rectangular region is considered as phrase pair. Repeating this step iteratively identifies all the phrase pairs. Equation (1) is used to calculate the point wise mutual information [17]:

$$I(e, f) = log_2 \frac{P(e,f)}{P(e)p(f)} \qquad (1)$$

After the segmentation of sentence pairs into phrase pairs, joint probabilities are calculated for these phrase pairs using monolingual conditional probability.

The advantage of this method is that it does not require to find high frequency N-grams. Precision and length penalty is used as measuring scores here and a confidence level of 99.99% over baseline HMM represents the efficiency of this approach.

### 2.7. Bilingual N-gram based approach

Ashish Venugopal, Stephan Vogel and Alex Waibel developed an approach for phrase translation extraction using N-grams and the method builds phrase lexicons from bilingual corpus [18]. The method consists of three phases: generation, scoring, and pruning. In the generation phase, all source phrases are identified using N-gram and all possible candidate target phrases corresponding to source phrases are identified. This set is scored and pruned using various scores to remove unwanted target phrases. In the scoring phase, the phrases are scored using measures from three models, maximum approximation, word based translation lexicon and language specific measures [18]. In the pruning step, maximum likely phrase pairs are selected using maximal separation criteria [18].

The advantage of this method is that it is less computationally expensive and recovers well from noisy alignments, but it lacks an explanatory framework. When performance is considered, the method shows a NIST score improvement of 0.05 over baseline HMM word alignment method.

## 2.8. Block based method

Ashish Venugopal, Stephan Vogel and Alex Waibel developed a phrase translation extraction from alignment models in 2005, which is based on blocks [21]. In the block based method [21], phrase pair within parallel sentence is considered as a block. The method does not use alignment. An example of a block is shown in fig.1 [21].

In the block, y-axis is the source sentence, x axis is the target sentence. The block is defined by source phrase and its projection, which can be represented as the left and right boundaries in the target sentence. The source phrase is bounded by the start and end positions in the source sentence.

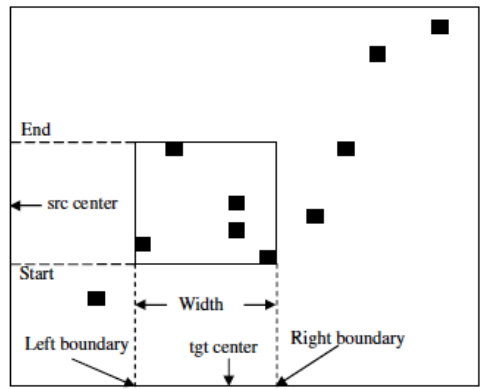

Fig. 1. Example of a block

Three models are used in this method. The first one is Fertility model, which predicts the width of the block by computing phrase length. A dynamic programming algorithm using the source word fertilities is employed in this model and given the candidate target phrase e and a source phrase of length J, the model gives the estimation of $P(J/e1)$. Next is the Distortion model. A simple distortion score is computed to estimate how far away the two centers are in a parallel sentence pair in a sense the block is close to the diagonal [21].Another model is Lexicon model, which is computed for translational equivalence. For each candidate block, using word level lexicon, a score within a given sentence pair is computed.

For each candidate block, the scores of phrase length, center based distortion and a lexicon-based score are calculated, which is followed by a local greedy search to find best scored phrase pair. The method is a general framework, in which one could plug in other scores and word alignment to get better results, but the computational expense of this method will be higher.

## 2.9. Clustering method

Rile HU, Chengqing ZONG and Bo XU proposed an approach to automatic acquisition of translation templates which is based on phrase structure extraction and alignment [23]. The method is a statistical and data driven approach [23]. The basic idea of the method is to cluster or group words in the corpus using similarity measure. Clustering is performed in two steps, temporal clustering and spatial clustering. Temporal clustering clusters words or entities, which frequently co occurs, into groups. Frequent co occurrence of entities can be obtained by finding the mutual information score between the word pair or entity pair. Spatial clustering clusters words or entities, which have similar left and right contexts which is measured by the Kullback-Leibler distance [23].

The initial step of the method is to find the similarity measure in terms of distance for each pair of entities and N-pairs of entities with minimum distance is selected to form semantic class. The entities are then replaced with semantic class label. The next step is to find similarity measure in terms of the Mutual Information for each entity pair and N pairs of entities with highest Mutual Information (MI) are selected to form phrasal structure groups. The entities are replaced with phrasal structure class label. This process is repeated till a stopping criterion (STC) is reached.

The method has higher precision, recall and f-score than base approaches Bracketing Transduction Grammar method [8] and parse to parse match[13]. Higher number of phrase groups were obtained when cosine of pointwise mutual information score was used [23].

### 2.10. Loose phrase extraction method

Xue Yongzeng and Li Sheng developed a loose phrase extraction method with n-best alignments [28]. The method of loose phrase extraction [28] is based on the idea of extracting phrase pairs that are not strictly consistent with word alignments. In normal phrase extraction techniques, which use bilingual corpora, exact phrase extraction is used, that is, all the phrase pairs that are consistent are extracted. But this method allows some relaxation to the rule of consistency. Loose phrase pairs can be aligned to some words outside, provided that the word is also aligned to some words inside the phrase pair [28]. After phrase extraction, constraints are applied to these loose phrase pairs to avoid ill formed phrase pairs. Main constraints applied are intersection-based constraint and heuristic based constraint. Apart from applying constraints, a union between n-best alignments from each translation direction is collected and the two unions of alignment are combined. The method achieves better performance than baseline exact match approach, also N-best alignment results on all constraints shows better BLEU score.

### 2.11. Word alignment and Rule based approach

Andreas Eisele, Christian Federmann, Hervé Saint Amand, Michael Jellinghaus, Teresa Herrmann and Yu Chen developed a hybrid method integrating a rule based with a hierarchical translation system [30]. This method is a statistical and rule based hybrid approach. The hybrid system inherits the lexicons from both sub-systems as well as other merits of each system [30]. The method uses word alignment method and rule based approach to extract phrases. Phrase tables are generated from both statistical method and rule based method. These phrase tables are later combined so that the hybrid system can exploit knowledge from both methods [30]. The advantage is that the hybrid method can gain extra knowledge from rule-based system but the errors in rule-based system can affect the correct information in statistical system [30].

Another method with variation in combining translation models from various sources has been proposed in 2010 by Yu Chen and Andreas Eisele [32]. In this method, instead of combining phrase tables by adding one binary feature for each individual system, all features in both translation models are retained while combining.

BLEU score is used as a measuring score. The hybrid method showed improved performance than the baseline word alignment or rule based approach. The performance difference between the hybrid system and the SMT core improved to nearly 1.5 BLEU [32].

### 2.12. N-gram and Rule based approach

N-gram and Rule based approach [34] is a statistical and rule based hybrid approach developed by Yoh Okuno. In this method, N-gram model is applied to preprocessed corpus using map reduce framework. The N-gram model application is followed by rule based filtering based on the part of speech patterns. Three types of errors were observed, Judgment inconsistency, Morphological analysis error, Lack of features for additional rules. Measuring scores used are precision, recall and F-measure. When compared with the baseline N-gram, precision improvement of about 0.49 was observed. The method shows better performance than N-gram and rule based approaches.

## 3. CLASSIFICATION

The level 1 classification of the different phrase structure learning methods is given in figure.2. The different methods of phrase structure learning can be broadly classified into three classes, statistical methods, rule-based methods and statistical and rule hybrid methods. The approaches, which use statistical techniques for phrase extraction are classified as statistical methods and include clustering method, mutual information based, N-gram based, syntactic based method, alignment based method, block based method. Statistical methods are based on the statistical modeling of data and depend on statistical theorems and rules. The methods do not require rule set for the language.

The approach, which uses basic rules for phrase structure extraction, is classified as rule based method. The rule based approach needs a developed set of rules for the language and the task is performed based on this set of rules. The approaches, which use both statistical techniques and rules for phrase structure extraction, are classified as statistical and rule hybrid methods and include N-gram and rule based method and word alignment and rule based method.

## 4. OBSERVATIONS AND DISCUSSION

The different methods for phrase structure learning can be compared in detail based on several factors. A level 2 classification of the methods based on corpus used, initial alignment requirement of corpus, base approach, tools used, and technique used is given in Table.1. Another level of classification, level 3 based on different scores applied, evaluation metrics used and efficiency is given in Table.2.

When the different methods are compared, an observation made is that the mutual information based method and probabilistic based method shows higher efficiency and performance. Mutual information based method shows a confidence value of 99.99% [17], which is promising. Hybrid methods like N-gram and rule based approach and word alignment and rule based approach shows good performance but requires set of rules for the language.

Clustering method also shows relatively better results but the concept of clustering and translation templates does not have much relevance when agglutinative languages like Indian languages, Japanese, Turkish etc are considered, and may need to consider words in its root form [10]. Alignment template approach, which is a word alignment based method, does not distinguish between good or bad templates though it can be learned automatically using bilingual corpus. When N-gram based approached are considered, basic N-gram shows less accuracy as internal structure of phrases are not considered. Bilingual N-grams show good performance than basic and is less computationally expensive, but it lacks an explanatory framework. Block based method is another candidate but it can be computationally expensive. Rule based approaches is not so efficient as the task of developing a rule set for a particular language is cumbersome and also large amount of rules have to be developed manually. Thus it needs time and support from trained linguistics expert. But when compared with N-grams, the accuracy is more for rule-based approach as it takes into consideration the internal structure of phrases. Hybrid methods like N-gram and rule based approach and word alignment and rule based approach shows good performance but requires set of rules for the language. Syntactic method is not much efficient, as syntactic models do not provide important phrase alignments. Weighting of syntactic phrases also does not improve the performance much.

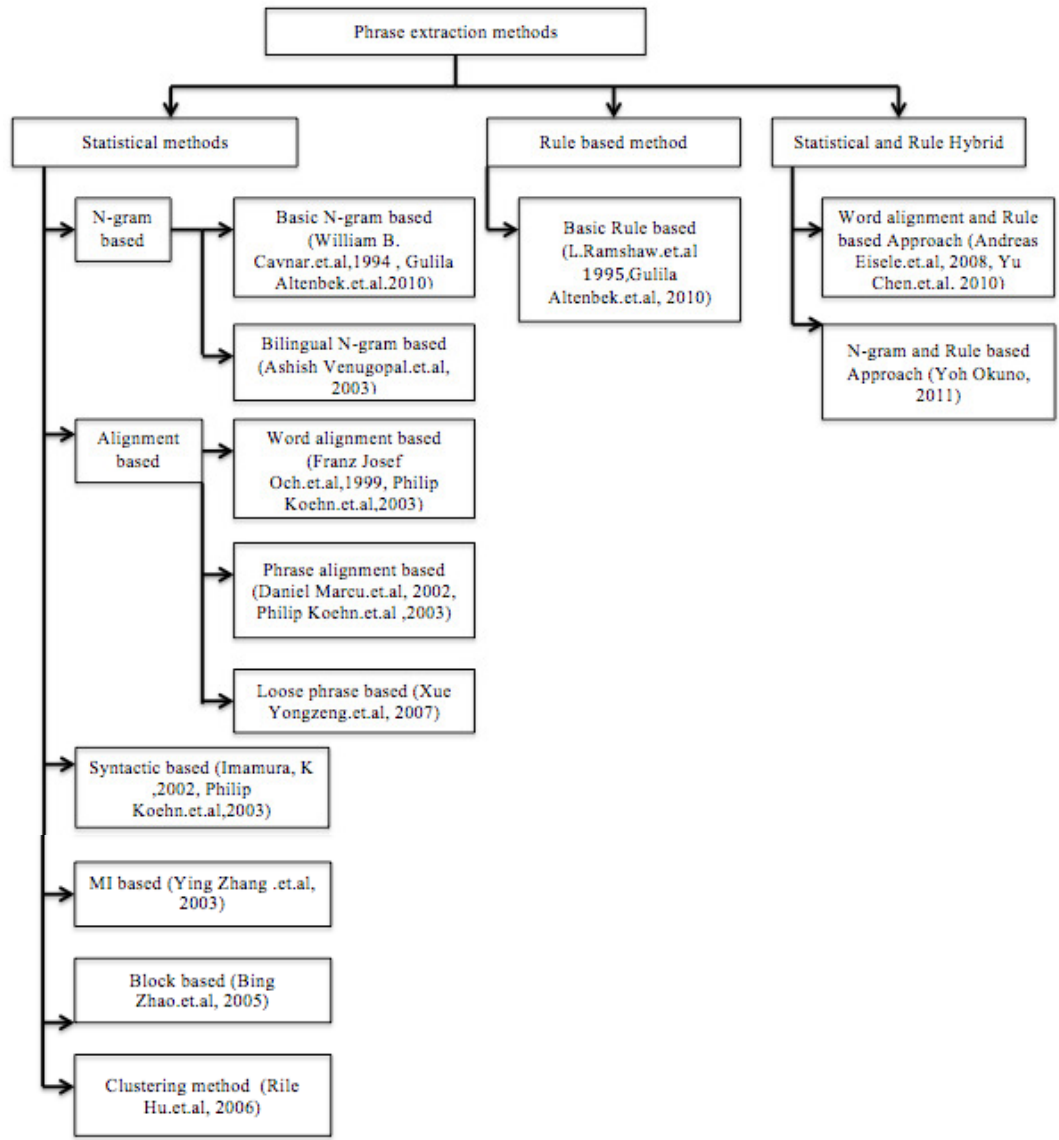

Figure.2. Level 1 classification of methods

| Method | Author and year | Corpus used | Initial alignment requirement of corpus | Base approach | Tools used | Technique used |
|---|---|---|---|---|---|---|
| Basic N-gram based method | William B. Cavnar.et.al (1994), Gulila Altenbek.et.al (2010) | Xinjiang daily corpus | No information available | No information available | No information available | N-gram method for unannotated text |
| Basic Rule based approach | L. Ramshaw.et.al (1995), Gulila Altenbek.et.al (2010) | Xinjiang daily corpus | No information available | No information available | No information available | Noun phrases extracted based on set of rules |
| Word Alignment based method | Franz Josef Och.et.al (1999), Philip Koehn.et.al (2003) | Europarl corpus | Sentence aligned | Single word based approach(uses manual dictionary) | GIZA++ tool kit | Phrases from word based alignments - alignment template approach and modified alignment template approach |
| Phrase alignment based method | Daniel Marcu.et.al (2002), Philip Koehn.et.al (2003) | Europarl corpus | Sentence aligned | word alignment method(IBM model 4) | No information available | Phrases are extracted from the phrase alignment using phrase-based joint probability model |
| Syntactic method | Imamura, K (2002), Philip Koehn.et.al (2003) | Europarl corpus | Sentence aligned | Syntactic translation models | Syntactic parser | Syntactic phrase pairs extracted from word aligned corpus, word alignment-modified alignment template approach |

| Method | Author and year | Corpus used | Initial alignment requirement of corpus | Base approach | Tools used | Technique used |
|---|---|---|---|---|---|---|
| Mutual Information based method | Ying Zhang .et.al (2003) | Xinhua English news corpora | Sentence aligned | Word-word translations (IBM model1), Phrase-phrase translations from HMM word alignment | No information available | Integrated phrase segmentation and alignment algorithm- phrases identified by similarity of point wise mutual information |
| Bilingual N-gram based method | Ashish Venugopal .et.al (2003) | English- Chinese parallel language corpus | Word aligned using IBM alignment model | HMM alignment model, word level system( IBM model1) | GIZA tool | Building phrasal lexicons by N-gram method with generation, scoring, pruning steps |
| Block based method | Bing Zhao.et.al (2005) | English- French corpus | Sentence aligned | word alignment method(IBM model 4) | GIZA++, pharaoh decoder | Fertility model-to predict width of the block. Distortion model- to predict how close centers of source and target phrase are. Lexicon model- for translation equivalence. |
| Clustering method | Rile Hu.et.al (2006) | English- Chinese parallel spoken language corpus | Sentence aligned | Phrase alignment method using Bracketing Transduction Grammar, Syntactic method using parse-to-parse match | No information available | Phrase extraction- temporal and spatial clustering, Alignment- bracketing transduction grammar |
| Loose phrase based method | Xue Yongzeng. et.al (2007) | IWSLT-04 Chinese- English translation task | Sentence aligned | Word alignment method using IBM word alignment model | Pharaoh trainer and decoder | Based on loose phrase extraction, with extensions of word position based constraints and n-best alignments |
| Word alignment and Rule hybrid | Andreas Eisele.et.al (2008), Yu Chen.et.al (2010) | Europarl corpus | Sentence aligned | Rule based approach and Word alignment based method(IBM word alignment model) | Moses or Joshua decoder, Lucy, SRILM toolkit | Based on word alignment method and rule based method |

| N-gram and Rule hybrid | Yoh Okuno (2011) | Japanese blog corpus | No information available | N-gram based method | No information available | Based on N-gram model and rule based filtering |

Table 1. Level 2 classification based on the technique and corpus used

| Method | Author and year | Scores applied | Evaluation metrics used | Efficiency |
|---|---|---|---|---|
| Basic N-gram based method | William B. Cavnar.et.al (1994), Gulila Altenbek.etal (2010) | No information available | Accuracy % measured in terms of number of phrases correct to total phrases extracted | Bigram accuracy- 54.1%, Trigram accuracy- 51.5% |
| Basic Rule based approach | L. Ramshaw.et.al (1995), Gulila Altenbek.etal (2010) | No information available | Accuracy % measured in terms of number of phrases correct to total phrases extracted | Accuracy- 80.3% |
| Word Alignment based method | Franz Josef Och.et.al (1999), Philip Koehn.et.al (2003) | lexical weight, word alignment heuristics | BLEU score | Error rate decreased to about 6%, improvement of about 0.01 BLUE score |
| Phrase alignment based method | Daniel Marcu.et.al (2002), Philip Koehn.et.al (2003) | t-counts | BLEU score | Improvement of 0.02 BLEU score |
| Syntactic method | Imamura, K (2002), Philip Koehn.et.al (2003) | syntactic phrase weight, lexical weight | BLEU score | BLEU: 0.243 |
| Mutual Information based method | Ying Zhang.et.al (2003) | student's t-test | NIST- precision and length penalty, Chinese English machine translation evaluation package | Confidence:99.99% Precision-6.966, Lengthpenality-0.97 |

| Method | Author and year | Scores applied | Evaluation metrics used | Efficiency |
|---|---|---|---|---|
| Bilingual N-gram based method | Ashish Venugopal .et.al (2003) | Maximum approximation, estimation from word based translation lexicon, language specific measure, score within sentence consistency, across sentence consistency, score on maximal separation criteria | NIST and BLEU scores | NIST improvement over HMM: 0.05 Scores: BLEU : 0.197 NIST: 7.6 |
| Block based method | Bing Zhao.et.al (2005) | Fertility probability, distortion center, word level lexicon probability, seven base scores | BLEU score | Dev.Bleu: 27.44 , Tst.Bleu: 27.65 |
| Clustering method | Rile Hu.et.al (2006) | Distance, Cosine measure, Cosine of pointwise MI, Dice Coefficient | Precision(P), Recall(R), F-measure(F) | P-76.7%, R-80.8%, F-78.75% |
| Method | Author and year | Scores applied | Evaluation metrics used | Efficiency |
| Loose phrase based method | Xue Yongzeng. et.al (2007) | Intersection based constraint, Heuristic based constraint | BLEU score | Improvement over baseline: 0.043 |
| Word alignment and Rule hybrid | Andreas Eisele.et.al (2008), Yu Chen.et.al (2010) | No information available | BLEU score | Moses+Lucy: 27.26 Joshua+Lucy: 27.52 |
| N-gram and Rule hybrid | Yoh Okuno (2011) | No information available | Precision, Recall and F-measure | P:0.9,R:0.81,F:0.85 |

Table 2. Level 3 classification based on scores applied and efficiency

## 5. CONCLUSION

Text classification is an important natural language processing task, which has got many useful applications like spam filtering, email routing, language identification, genre classification, readability assessment. The use of phrases helps in capturing non-local behaviors and thus helps in the improvement of text classification task. In this survey paper, different phrase structure learning methods for text classification have been studied. The approaches are classified in a broader aspect into three groups, statistical methods, rule-based methods and statistical and rule hybrid methods. Different techniques are further classified into two more levels based on the technique used and efficiency. The methods are classified and compared in detail based on different factors like corpus used, initial alignment requirement, base approach, tools used, techniques in the level 2 classification. The methods are again compared and classified based on different scores applied, evaluation metrics used and efficiency in level 3 classification. One observation made is that the major works in phrase structure learning are focused on statistical and hybrid methods as rule based approach needs time and trained linguistics personnel. The major observation made is that mutual information based approach is the most promising technique for phrase structure extraction and shows better performance and efficiency.

**Authors**

**Reshma Prasad** received her B.Tech degree in Computer Science & Engineering from Amrita School of Engineering, Kollam in 2010. She is currently doing Mtech in Computer Science and Information Systems from Rajagiri School of Engg. & Technology. Her research interests include Natural Language Processing and Text classification.

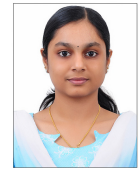

**Mary Priya Sebastian** received her B.Tech degree in Computer Engineering from College of Engineering, Chengannur in 2002 and M.Tech degree in Computer and Information Science from the Department of Computer Science, CUSAT in 2010. She is currently working as Assistant Professor in the Department of Computer Science, RSET. She is currently doing her Ph.D in the Dept. of Computer Science, CUSAT. Her research area is Natural Language Processing and Machine Learning. She has 12 publications including National and International ones and she is a professional member of ACM.

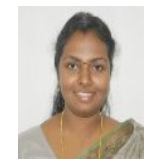